\documentclass[11pt,a4paper]{article}
\usepackage[hyperref]{acl2021}
\usepackage{times}
\usepackage{adjustbox}
\usepackage{latexsym}
\usepackage{verbatim}

\PassOptionsToPackage{linktocpage}{hyperref}

\renewcommand{\UrlFont}{\ttfamily\small}
\newcommand{\B}{\bfseries}

\usepackage{microtype}

\aclfinalcopy 
\def\aclpaperid{***} 

\title{Sentence Embeddings by Ensemble Distillation}

\author{Fredrik Carlsson \\
 RISE \\
 Sweden \\
 \texttt{fredrik.carlsson@ri.se} \\\And
 Magnus Sahlgren \\
 RISE \\
 Sweden \\
 \texttt{magnus.sahlgren@ri.se} \\}

\date{}

\begin{document}
\maketitle
\begin{abstract}
This paper contributes a new State Of The Art (SOTA) for Semantic Textual Similarity (STS). 
We compare and combine a number of recently proposed sentence embedding methods for STS, and propose a novel and simple ensemble knowledge distillation scheme that improves on previous approaches. Our experiments demonstrate that a model trained to learn the average embedding space from multiple ensemble students outperforms all the other individual models with high robustness. Utilizing our distillation method in combination with previous methods, we significantly improve on the SOTA unsupervised STS, and by proper hyperparameter tuning of previous methods we improve the supervised SOTA scores.
\end{abstract}

\section{A Note on Terminology}
Common practice in work on STS is to use the term ``unsupervised'' to refer to not explicitly utilizing labeled STS data. Approaches using other sorts of labeled data, such as the NLI data for S-BERT, is therefore described as unsupervised in this context. This is somewhat confusing, and not in line with our interpretation of the term unsupervised. Hence, throughout this paper we will use the 
term ``zero-shot'' when referring to algorithms that do not explicitly use labeled STS data, reserving the term ``unsupervised'' for methods which are {\em completely} unsupervised/self-supervised throughout their training process. We hope proceeding work will follow this convention in order to align the terminology with other fields of Machine Learning and NLP.

\section{Introduction}
Creating semantically coherent sentence representations is an important and long standing goal in NLP. The use of such sentence representations is ubiquitous, and include tasks such as sentiment analysis \cite{yu-jiang-2016-learning}, question answering \cite{yadav-etal-2019-alignment}, machine translation \cite{wang-etal-2017-sentence}, and natural language inference \cite{talman}.
The quality of sentence representations is measured either by their performance in downstream tasks, or by their correspondence to human semantic similarity judgements \cite{conneau2018senteval}. This latter task is referred to as Semantic Textual Similarity (STS), and has become the standard way to benchmark the quality of sentence representations.

This paper investigates the combination of recent SOTA methods for unsupervised and zero-shot STS. 
In addition to this, we introduce a simple, but novel, method for knowledge distillation when the aim is to learn sentence embeddings. Our method relies on the realisation that averaging the actual embeddings of multiple teacher models makes for good student regression targets. We evaluate our method on STS and find our method to be robust and work well with already established methods.

Additionally, we observe that previous supervised embedding methods are hampered by poor hyperparameter tuning. By performing a simple grid-search over the possible relative regression targets proposed by \citet{carlsson2021semantic}, certain models see a significant performance increase. This results in a new SOTA for supervised STS.


Our top performing models are available at: \url{https://github.com/FreddeFrallan/Contrastive-Tension}

\section{Related Work}
\subsection{Sentence Embeddings}
Work on sentence embeddings can be divided into three different categories. The arguably simplest approach (except, of course, bag-of-words) is to rely on composition of pre-trained word embeddings \cite{le2014distributed, wieting2015towards, arora2016simple, DBLP:conf/iclr/WietingK19}. Another approach is to rely on some form of supervision, which often uses an initial supervised proxy task before potentially training directly for STS. Examples of such supervised approaches include InferSent by \citet{conneau-etal-2017-supervised} which learns sentence embeddings via a siamese BiLSTM trained on NLI data. \citet{cer-etal-2018-universal} propose the Universal Sentence Encoder (USE), which is a Transformer encoder trained with both unlabeled data and labeled NLI data. Finally, S-BERT by \citet{SBert} adopts the training objective of InferSent but instead applies it to pre-trained BERT models \citep{BERT}.

A third category of sentence embedding methods is self-supervised methods that require no labeled data. Examples of such methods include Skip-Thoughts of \citet{NIPS2015_5950}, which trains an encoder-decoder to reconstruct surrounding sentences given an encoded passage. \citet{QuickThoughts} propose QuickThoughts, which instead formulates the training objective as a sentence context classification task. \citet{BertFlow} propose to use Normalizing Flows to transform a pre-trained model's embedding space into a Gaussian Distribution. Finally, \citet{carlsson2021semantic} introduce Contrastive Tension, which tunes two independent models to yield similar embeddings for identical sentences, and dissimilar embeddings for non-identical sentences.

This paper investigates in particular the combinations of the supervised Natural Language Inference (NLI) task of \citet{SBert}, the unsupervised Gaussian Flow Normalization (Flow) of \citet{BertFlow}, and the Self-Supervised Contrastive Tension (CT) algorithm of \citet{carlsson2021semantic}. 

\subsection{Knowledge Distillation}
Knowledge distillation is a method where instead of training a model directly towards a specific task, it is trained towards the information of either a strong teacher model, or an ensemble of teacher models \cite{HintonDistillation}.
\citet{SentenceEmbeddingDistillation} use this technique to learn an embedding space by training it to contain the same pairwise relationships between sentences as that of a fine-tuned Transformer model fine-tuned on sentence-pair tasks. 
\citet{MultilingualSentenceDistillation} distill cross-lingual sentence embeddings using parallel corpora, by directly approximating the embedding space of a English S-BERT fine-tuned on labeled STS data.

Our distillation scheme is mostly similar to that of \citet{MultilingualSentenceDistillation}, in the sense that we directly train a model's embedding space to approximate an already existing space. However, we distill from an ensemble of models within the same language, by averaging the ensemble model's embedding space.

\section{Sentence Ensemble Distillation}
Our Sentence Ensemble Distillation (SED) objective aims to train a new model, which we refer to as ``the distillation learner'', by distilling the information from multiple already existing sentence embedding models. This is realised by training the distillation learner to minimize the Mean Squared Error (MSE) between its embeddings and the average sentence embedding from that of the ensemble models. Our SED objective only updates the weights of the distillation learner.

Formally, we describe our SED objective in the following terms. Using a set of $N$ sentence embedding models $E = \{M_0, \ M_1, \ M_2, ... M_N\}$, where $M_i(S)$ produces a vector with $D$ dimensions, given sentence $S$. The mean ensemble embedding $S_E$ is calculated as:  
\begin{equation}
S_E = \frac{1}{N} \sum^N_{i=1} M_i(S).     
\end{equation}

The per-example loss for distillation learner $f$ is then calculated as:
\begin{equation}
J(S) = \textrm{MSE}(S_E, \ f(S)) 
\end{equation}
\noindent
where $f(S)$ also produces a vector with $D$ dimensions. 

\section{Experiments}
\label{Experiments-main}

\begin{table*}[!htbp]
\centering
\caption{Pearson and Spearman correlation (x100) for Unsupervised algorithms on various zero-shot semantic textual similarity tasks. \textbf{*} indicates reported results that are taken directly from their original paper. }
\begin{adjustbox}{width=1\textwidth}
\begin{tabular}{ |l|c|c|c|c|c|c| } 
 \hline
  & STS12 & STS13 & STS14 & STS15 & STS16 & Avg. \\ 
 \hline
Avg GloVe   \textbf{*} & 52.30 / 53.30  & 50.50 / 50.70 & 55.20 / 55.60 & 56.70 / 59.20 & 54.90 / 57.70  & 53.92 / 55.3 \\
 Skip-Thoughts \textbf{*} & 41.00 / -  & 29.80 / - & 40.00 / - & 46.00 / - & 52.00 / -  & 41.76 / - \\
\hline
 BERT-L & 42.59 / 49.01 & 47.35 / 50.88 & 49.31 / 49.69 & 55.56 / 56.79 & 60.43 / 61.41 & 51.05 / 53.56\\
 BERT-L-Flow & 58.94 / 60.35 & 65.34 / 65.96 & 64.67 / 62.09 & 71.19 / 71.88 & 72.27 / 71.88 & 66.48 / 66.43 \\
 BERT-L-Flow \textbf{*} & - / 65.20 & - / 73.39 & - / 69.42 & - / 74.92 & - / 77.63 & - / 72.11 \\
 BERT-L-CT &  69.63 / 69,50 & 75,79 / 75,97 & 77,15 / 74,22 & 78,28 / 78,83 & 77,70 / 78,92 & 75,71 / 75,49 \\ 
 \hline
 \textit{\textbf{Our Contributions}} & & & & & & \\
   \hline
 BERT-L-CT-Flow & 69.78 / 69.51 & 77.32 / 77.49 & 77.71 / 74.53 & 80.32 / 80.73 & 79.46 / 79.56 & 76.92 / 76.36 \\
 BERT-L-CT-SED & 69.74 / \B 69.87 & 78.12 / 77.84 & 78.71 / \B 75.87 & 79.43 / 80.48 & 78.76 / \B 80.74 & 76.95 / 76.96 \\
 BERT-L-CT-SED-Flow & \textbf{70.15} / 69.76 & \B 78.75 / 78.91 & \textbf{78.73} / 75.53 & \B 81.15 / 81.39 & \textbf{79.96} / 80.14 & \B 77.75 / 77.15 \\
 \hline
\end{tabular}
\end{adjustbox}
\label{Unsupervised STS Table}
\end{table*}

\subsection{Setup}
Both distillation learners and ensemble models are pre-trained Transformers and always share the same architecture. For brevity, we limit this to BERT-Large, with the exception of the Supervised STS tasks where we also include RoBERTa-Large \citep{RoBerta}. Additional results for BERT-Distill and BERT-Base are available in Appendix-\ref{bert-base}.
During SED we generate sentence embeddings from both the ensemble and the distillation learner by mean pooling the final layer. When evaluating the distillation learner, we mean pool over the two final layers (See Appendix-\ref{LayerPoolingAblation} for a layer-pooling study).

\begin{table*}[!htbp]
\centering
\caption{Pearson and Spearman correlation (x100) for algorithms relying on labeled NLI data on various zero-shot semantic textual similarity tasks. \textbf{*} indicates reported results that are taken directly from their original paper. }
\begin{adjustbox}{width=1\textwidth}
\begin{tabular}{ |l|c|c|c|c|c|c| } 
 \hline
  & STS12 & STS13 & STS14 & STS15 & STS16 & Avg. \\ 
 \hline
 Infer-Sent \textbf{*} & 59.30 / 60.30 & 58.80 / 58.70 & 69.60 / 66.70 & 71.30 / 72.20 & 71.50 / 72.60 & 66.10 / 66.10 \\
 BERT-L-NLI & 67.74 / 67.48 & 70.93 / 72.61 & 75.3 / 75.26 & 78.15 / 79.18 & 72.98 / 76.09 & 73.02 / 74.12 \\
 BERT-L-NLI-CT & 72.7 / 71.07 & 78.01 / 78.17 & 80.42 / 78.23 & 81.4 / 82.34 & 78.29 / 80.01 & 78.16 / 77.96 \\
 BERT-L-NLI-Flow & 71.0 / 69.73 & 75.53 / 75.59 & 79.29 / 77.11 & 81.8 / 81.8 & 78.53 / 78.79 & 77.23 / 76.6 \\
 BERT-L-NLI-Flow \textbf{*} & - / 70.19 & - / \B 80.27 & - / \B 78.85 & - / 82.97 & - / \B 80.57 & - / \B 78.57 \\
 \hline
 \textit{\textbf{Our Contributions}} & & & & & & \\
   \hline
 BERT-L-NLI-CT-Flow & \textbf{72.76} / 70.89 & 76.75 / 76.82 & \textbf{80.55} / 77.85 & \B 82.9 / 83.12 & \textbf{79.98} / 80.18 & 78.59 / 77.77 \\
 BERT-L-NLI-CT-SED & 71.87 / 70.94 & \textbf{78.91} / 78.81 & 79.98 / 78.09 & 81.35 / 82.77 & 76.42 / 78.94 & 77.71 / 77.91 \\
 BERT-L-NLI-CT-SED-Flow & 72.56 / \textbf{71.55} & 78.70 / 79.29 & 80.88 / 78.06 & 82.71 / 82.98 & 79.27 / 79.72 & \textbf{78.82} / 78.32 \\
 \hline
\end{tabular}
\end{adjustbox}
\label{Zero-Shot STS Table}
\end{table*}

The fine-tuning objectives and their combinations are applied to each ensemble model sequentially, where the model name describes the order. However, the Flow normalization is only applied to the final distillation learner, and not to the ensemble models. For example, BERT-L-NLI-CT-SED-Flow is a Flow normalized BERT-Large model trained with SED from an ensemble of BERT-Large models, which each have been first fine-tuned towards the siamese NLI task and then tuned with CT.

All SED experiments use the same set of hyperparameters, with the same set of $100k$ randomly sampled sentences from the English Wikipedia and always using an ensemble of $10$ models. The distillation is performed with the Adam optimizer using a batch size of $32$, a final learning rate of $2e^{-5}$ and a linear warm-up schedule for $10\%$ of the training data. The hyperparameters for the separate fine-tuning tasks of NLI, CT and Flow are taken from their respective publications, and are listed in Appendix-\ref{Hyperparameters} for completeness.

\subsection{Unsupervised STS}
\label{UnsupervisedExperiments}
Table \ref{Unsupervised STS Table} shows results for the various unsupervised methods and their combinations when evaluated on the zero-shot English STS tasks of \citet{agirre-etal-2012-semeval, agirre-etal-2013-sem, agirre-etal-2014-semeval, agirre-etal-2015-semeval, agirre-etal-2016-semeval}.
As the SED itself requires no labeled data, the distillation learner is deemed unsupervised if its ensemble is trained without labeled data. Hence, we utilize both Flow and CT for these ensemble models, but not the siamese NLI objective.

We observe that the combination of previously proposed methods lead to improved performance, and that adding the SED method further increases the performance. We set a new SOTA for unsupervised STS with an average score of 77.75 (Pearson) and 77.15 (Spearman). We also note a sometimes significant discrepancy between previously reported results and results replicated in our experiments.

\subsection{Zero-Shot STS}
\label{Zero-ShotExperiments}
Table \ref{Zero-Shot STS Table} shows results for the various methods which require labeled data when evaluated on the English zero-shot STS tasks. We observe a similiar tendency as in section-\ref{UnsupervisedExperiments}, combining previous approaches leads to improved performance. However, the addition of SED does not contribute as much for the zero-shot methods. Again, we note discrepancies between previously reported results and our own.

\subsection{Distillation Stability}
In order to investigate the stability of the SED method, we train $10$ distillation learners for each ensemble and compare their final performance on the zero-shot STS tasks. Table \ref{Stability} reports the max, mean and standard deviation for the average Spearman score on the zero-shot STS task, for each model in the different ensembles, and for their respective $10$ distillation learners. Finally, we report the results attained when utilizing the full ensemble for prediction.

The average case of the SED models all outperform the best performing model of their respective ensemble.  The standard deviation between the performance of different runs of SED is significantly smaller than that of the models within the ensembles.

\begin{table}[!htbp]
\centering
\caption{Max, mean and the standard deviation of the Spearman correlation (x100) for the average zero-shot STS tasks.}
\begin{adjustbox}{width=0.47\textwidth}
\begin{tabular}{ |l|c|c|c|} 
 \hline
  & Max & Mean & STD \\ 
  \hline
  \textbf{Ensemble Models} & & & \\
 \hline
 BERT-L-CT & 75.22 & 74.07 & 0.80  \\     
 BERT-L-CT-Flow & 75.30 & 74.41 & 0.48  \\
 BERT-L-NLI-CT & 77.35 & 76.91 & 0.32  \\
 BERT-L-NLI-CT-Flow & 77.87 & 77.50 & 0.25  \\
 \hline
   \textbf{Full Ensemble} &  & & \\
 \hline
 BERT-L-CT & 77.40 & - & - \\
 BERT-L-CT-Flow & 77.62 & - & -  \\
 BERT-L-NLI-CT & 78.06 & - & - \\
 BERT-L-NLI-CT-Flow & 77.91 & - & -  \\
  \hline
  \textbf{Distillation Learners} & & &   \\
  \hline
 BERT-L-CT-SED & 77.04 & 76.76 & 0.20 \\
 BERT-L-CT-SED-Flow & 77.15 & 76.73 & 0.26  \\
 BERT-L-NLI-CT-SED & 78.08 & 77.93 & 0.09 \\
 BERT-L-NLI-CT-SED-Flow & \B 78.32 & \B 78.19 & 0.10  \\
 \hline
\end{tabular}
\end{adjustbox}
\label{Stability}
\end{table}

\subsection{Supervised STS}
Building upon the supervised STS regression task proposed by \citet{SBert}, we propose to set the target upper bound to $1$ and find the lower bound for each model by training $10$ models for all $20$ lower bounds in $\{0, \ 0.05, \ 0.1, ... \ 0.95\}$, comparing the performance on the STS-b development set. Using the top performing lower bound, we create an ensemble of $10$ supervised STS models. For each ensemble we then use SED with early stopping in regards to the STS-b development set. 

Similar to the findings of \citet{carlsson2021semantic}, who found no improvements to the supervised task with either NLI or CT, we see no improvements in Table~\ref{Supervised STS Table} of either SED or Flow. However, the grid hyperparameter search over the regression lower bounds results in significant improvements for RoBERTa-large, while having no noticeable effect on BERT-Large. The improvements seen with RoBERTa-Large result in a new supervised STS SOTA, of 87.56 (Pearson) and 88.50 (Spearman).

\begin{table}[!htbp]
\centering
\caption{Pearson and Spearman correlation (x100) for Supervised STS on the STS-b test set. (1) indicates the pooling is done from the final layer.  \textbf{*} indicates results taken directly from their original paper. }
\begin{adjustbox}{width=0.47\textwidth}
\begin{tabular}{ |l|c|} 
 \hline
  & STS-b Test \\ 
 \hline
 BERT-L (1) & 85.90 / 86.35  \\
 RoBERTa-L (1) \textbf{*} & - / 86.39  \\
   \hline
  \textbf{Our Contributions - No STS Data} &  \\
 \hline
  BERT-L-CT-SED-Flow & 80.98 - 79.55 \\
  BERT-L-NLI-CT-SED-Flow & 83.14 / 82.05 \\
  \hline
  \textbf{Our Contributions - With STS Data} &  \\
  \hline
 BERT-L & 85.40 / 86.35  \\
 RoBERTa-L  & \B 87.56 / 88.50 \\
 \hline
 BERT-L-Flow  & 83.85 / 83.32  \\
 RoBERTa-L-Flow  & 86.85 / 86.52  \\
 BERT-L-SED & 85.11 / 85.80  \\
 BERT-L-SED-Flow & 83.15 / 82.30  \\
 RoBERTa-L-SED  & 87.15 / 88.29  \\
 RoBERTa-L-SED-Flow  & 86.29 / 85.92  \\
 \hline
\end{tabular}
\end{adjustbox}
\label{Supervised STS Table}
\end{table}



\section{Conclusion}
There have been several recent suggestions on how to improve the unsupervised ability of Transformer language models to quantify semantic textual similarity.
This paper analyzed the possibility to combine several of these proposed methods, and introduced a Sentence Ensemble Distillation (SED) method that trains a distillation learner to approximate the average embedding space of multiple existing embedding models. Combining previously proposed methods and our distillation objective, we set a new SOTA for unsupervised STS with an average of circa 2 Spearman points. We also contribute a new Supervised SOTA (with an improvement of circa 2 Spearman points) by proper hyperparameter tuning. 

Interestingly, we see no improvement for supervised STS when incorporating either previous methods or the SED objective. However, we note that the difference between the top unsupervised and the top supervised methods is only 6.45 (Spearman), and that the difference between our best fully unsupervised method and our best method using NLI data is only 1.17 (Spearman) on the zero-shot task. 

\bibliographystyle{acl_natbib}
\bibliography{anthology,acl2021, CustomRefs}

\clearpage

\appendix


\section{Additional Exeriments}

\subsection{Results with BERT-Distill \& BERT-Base}
\label{bert-base}
We perform identical experiments to the unsupervised STS in section-\ref{UnsupervisedExperiments} and the zero-shot STS experiments of section-\ref{Zero-ShotExperiments}, but also including BERT-Distill and BERT-Base. Table-\ref{UnsupervisedAllModels} shows the results for the methods which don't require any labeled data, and Table-\ref{ZeroShotAllModels} shows the results for methods relying on labeled NLI data.

In the unsupervised STS tasks we see that BERT-Base performs the worse out of the three models, with BERT-Large performing 2.54 Spearman points better than BERT-Distill. The difference between the three models are smaller when using NLI data, where BERT-Large is the best, followed by BERT-Base and finally BERT-Distill. The difference between the best BERT-Large model and the BERT-Distill model is less than 2 Spearman points. Interestingly, both BERT-Base and BERT-Distill react negatively to being Flow Normalized after SED, and their best combinations seems to be NLI-CT-SED.

\begin{table*}[!htbp]
\centering
\caption{Pearson and Spearman correlation (x100) for Unsupervised algorithms on various zero-Shot semantic textual similarity tasks.}
\begin{adjustbox}{width=1\textwidth}
\begin{tabular}{ |l|c|c|c|c|c|c| } 
 \hline
  & STS12 & STS13 & STS14 & STS15 & STS16 & Avg. \\ 
  \hline
 BERT-D-CT & 67.27 / 66.92 & 71.31 / 72.41 & 75.68 / 72.72 & 77.73 / 78.26 & 77.17 / 78.60 & 73.83 / 73.78\\ 
 BERT-B-CT & 67.19 / 66.86 & 70.77 / 70.91 & 75.64 / 72.37 & 77.86 / 78.55 & 76.65 / 77.78 & 73.62 / 73.29 \\ 
  BERT-L-CT &  69.63 / 69,50 & 75,79 / 75,97 & 77,15 / 74,22 & 78,28 / 78,83 & 77,70 / 78,92 & 75,71 / 75,49 \\ 
\hline
  \textbf{Our Contributions} & & & & & & \\
  \hline
  BERT Distill (xxx parameters) & & & & & & \\
  \hline
BERT-D-CT-Flow & 67.61 / 66.69 & 72.62 / 73.07 & 76.22 / 72.97 & 78.71 / 79.09 & 78.36 / 78.73 & 74.7 / 74.11 \\
BERT-D-CT-SED & 69.05 / 68.17 & 73.84 / 74.81 & 76.74 / 74.21 & 77.71 / 78.61 & 76.92 / 78.77 & 74.85 / 74.91 \\
BERT-D-CT-SED-Flow & 68.36 / 67.29 & 74.18 / 74.84 & 76.13 / 73.0 & 78.61 / 78.92 & 78.67 / 79.02 & 75.19 / 74.61 \\
 \hline
  BERT Base (xxx parameters) & & & & & & \\
  \hline
BERT-B-CT-Flow & 66.63 / 66.13 & 70.99 / 71.28 & 74.86 / 72.26 & 77.92 / 78.59 & 77.85 / 77.95 & 73.65 / 73.24 \\
BERT-B-CT-SED & 65.18 / 65.45 & 70.72 / 71.15 & 73.27 / 70.54 & 77.35 / 77.9 & 76.34 / 77.23 & 72.57 / 72.45 \\
BERT-B-CT-SED-Flow  & 66.54 / 65.92 & 72.36 / 72.15 & 73.69 / 70.96 & 77.78 / 78.22 & 77.8 / 78.09 & 73.63 / 73.07 \\
 \hline
 BERT Large (xxx parameters) & & & & & & \\
 \hline
  BERT-L-CT-Flow & 69.78 / 69.51 & 77.32 / 77.49 & 77.71 / 74.53 & 80.32 / 80.73 & 79.46 / 79.56 & 76.92 / 76.36 \\
 BERT-L-CT-SED & 69.74 / \B 69.87 & 78.12 / 77.84 & 78.71 / \B 75.87 & 79.43 / 80.48 & 78.76 / \B 80.74 & 76.95 / 76.96 \\
 BERT-L-CT-SED-Flow & \textbf{70.15} / 69.76 & \B 78.75 / 78.91 & \textbf{78.73} / 75.53 & \B 81.15 / 81.39 & \textbf{79.96} / 80.14 & \B 77.75 / 77.15 \\
 \hline
\end{tabular}
\end{adjustbox}
\label{UnsupervisedAllModels}
\end{table*}

\begin{table*}[!htbp]
\centering
\caption{Pearson and Spearman correlation (x100) for Unsupervised algorithms on various zero-Shot semantic textual similarity tasks.}
\begin{adjustbox}{width=1\textwidth}
\begin{tabular}{ |l|c|c|c|c|c|c| } 
 \hline
  & STS12 & STS13 & STS14 & STS15 & STS16 & Avg. \\ 
  \hline
   BERT-D-NLI-CT & 69.39 / 68.38 & 74.83 / 75.15 & 78.04 / 75.94 & 78.98 / 80.06 & 74.91 / 77.57 & 75.23 / 75.42 \\ 
 BERT-B-NLI-CT & 68.58 / 68.80 & 73.61 / 74.58 & 78.15 / 76.62 & 78.60 / 79.72 & 75.01 / 77.14 & 74.79 / 75.37 \\ 
  BERT-L-NLI-CT & 72.7 / 71.07 & 78.01 / 78.17 & 80.42 / 78.23 & 81.4 / 82.34 & 78.29 / 80.01 & 78.16 / 77.96 \\
  \hline
  \textbf{Our Contributions} & & & & & & \\
  \hline
  BERT Distill (xxx parameters) & & & & & & \\
 \hline
 BERT-D-NLI-CT-Flow & 69.81 / 68.34 & 75.58 / 75.6 & 79.55 / 76.91 & 81.96 / 82.33 & 78.99 / 79.24 & 77.18 / 76.48 \\
 BERT-D-NLI-CT-SED  & 72.02 / 70.12 & 76.91 / 77.68 & 79.98 / \B 78.41 & 79.68 / 81.11 & 76.74 / 79.52 & 77.07 / 77.37 \\
 BERT-D-NLI-CT-SED-Flow & 70.53 / 68.61 & 74.86 / 75.48 & 79.98 / 77.43 & 81.95 / 82.27 & 79.24 / 79.46 & 77.31 / 76.65 \\
 \hline
  BERT Base (xxx parameters) & & & & & & \\
  \hline
 BERT-B-NLI-CT-Flow & 70.48 / 69.16 & 75.32 / 75.83 & 79.96 / 77.37 & 82.05 / 82.07 & 79.0 / 79.16 & 77.36 / 76.72 \\
 BERT-B-NLI-CT-SED & 71.58 /  70.48 & 77.1 / 77.49 & 79.8 / 78.04 & 80.74 / 81.79 & 76.09 / 78.36 & 77.06 / 77.23 \\
 BERT-B-NLI-CT-SED-Flow & 69.69 / 68.84 & 75.77 / 76.25 & 79.75 / 77.19 & 82.6 / 82.58 & 78.89 / 79.16 & 77.34 / 76.8 \\
 \hline
  BERT Large (xxx parameters) & & & & & & \\
 \hline
  BERT-L-NLI-CT-Flow & \textbf{72.76} / 70.89 & 76.75 / 76.82 & \textbf{80.55} / 77.85 & \B 82.9 / 83.12 & \B 79.98 / 80.18 & 78.59 / 77.77 \\
 BERT-L-NLI-CT-SED & 71.87 / 70.94 & \textbf{78.91} / 78.81 & 79.98 / 78.09 & 81.35 / 82.77 & 76.42 / 78.94 & 77.71 / 77.91 \\
 BERT-L-NLI-CT-SED-Flow & 72.56 / \textbf{71.55} & 78.70 / \B 79.29 & 80.88 / 78.06 & 82.71 / 82.98 & 79.27 / 79.72 & \B 78.82 / 78.32 \\
 \hline
\end{tabular}
\end{adjustbox}
\label{ZeroShotAllModels}
\end{table*}

\subsection{SentEval}
\label{Sent Eval}

Although the SentEval test bed of \citet{SentEval}, has fallen a bit out of fashion as recent publications have been more focused towards STS, Table-\ref{Downstream} includes results for all its downstream tasks which are not the zero-shot STS tasks. The combination of NLI-CT-SED performs the best in most tasks and also in the average score. However, this improvement is quite marginal for most tasks. We find the Flow normalization to hamper performance on all tasks, sometimes significantly.

\begin{table*}[!htbp]
\centering
\caption{Results on the downstream tasks supplied with the SentEval package. For the semantic related tasks SICK-R and STS-b, the Pearson correlation (x100) is reported. }

\begin{adjustbox}{width=1\textwidth}
\begin{tabular}{ |l|r|r|r|r|r|r|r|r|r|r|r|r| } 
 \hline
  & CR & MR & MPQA & SUBJ & SST2 & SST5 & TREC & MRPC & SICK-E & SICK-R & STS-b & AVG \\ 
 \hline
 BERT-Large & 88.74 & 84.33 & 86.64 & \B 95.27 & 79.29 & 50.32 & \B 91.40 & 71.65 & 75.28 & 77.09 & 66.22 & 78.75 \\
 BERT-Large-NLI & \B 90.52 & 84.36 & 90.30 & 94.32 & 90.72 & 50.05 & 86.80 & \B 76.52 & 83.05 & 84.94 & 75.02 & 82.42 \\
 BERT-Large-CT & 86.81 & 82.38 & 88.31 & 94.34 & 87.75 & 46.56 & 88.00 & 73.10 & 81.49 & 84.93 & 76.50 & 80.92 \\
BERT-Large-NLI-CT & 89.56 & 82.56 & 90.20 & 83.08 & 88.85 & 48.60 & 87.20 & 74.43 & 82.77 & 84.88 & 77.49 & 80.87 \\
\hline
 \textit{\B Our Contributions} & & & & & & & & & & & & \\
 \hline
BERT-L-CT-SED & 88.82 & 83.62 & 89.04 & 95.23 & 89.18 & 49.77 & 91.00 & 74.32 & 82.52 & 85.82 & 77.38 & 82.43  \\
BERT-L-CT-SED-Flow & 88.79 & 83.17 & 88.79 & 95.04 & 88.54 & 46.56 & 91.00 & 63.30 & 77.43 & 79.80 & 66.13 & 78.96  \\
BERT-L-NLI-CT-SED & 90.01 & \B 84.46 & \B 90.61 & 95.02 & \B 90.77 & \B 51.09 & 90.20 & 74.38 & \B 83.34 & \B 86.60 & \B 79.43 & \B 83.28  \\
BERT-L-NLI-CT-SED-Flow & 89.93 & 83.70 & 90.40 & 94.86 & 90.17 & 49.77 & 89.80 & 71.13 & 79.72 & 81.68 & 72.83 & 81.64  \\
\hline

\end{tabular}
\end{adjustbox}
\label{Downstream}
\end{table*}

\subsection{Layer Pooling Ablation}
\label{LayerPoolingAblation}

The results in Table-\ref{LayerPoolTable} shows an layer-pooling ablation study for mean zero-shot STS Spearman correlation. Pooling from the two final layers is almost always preferable to only using a single layer. This goes for both fine-tuning objectives using NLI data and the combinations of self-supervised objectives. However, the differences between two and three layers are more sporadic. The best self-supervised method is found using two layers, and the best NLI score is found pooling from three layers.

\begin{table}[!htbp]
\centering
\caption{The average Spearman correlation (x100) for the zero-shot STS tasks, comparing three different pooling schemes.}
\begin{adjustbox}{width=0.47\textwidth}
\begin{tabular}{ |l|c|c|c|} 
 \hline
 Number of Pooling Layers & 1 & 2 & 3 \\ 
  \hline
  \textbf{Unsupervised} & & & \\
 \hline
 BERT-L & 53.56 & 54.76 & 55.52 \\
 BERT-L-Flow & 66.11 & 66.56 & 66.86 \\
 BERT-L-CT & 74.76 & 75.22 & 75.26  \\     
 BERT-L-CT-Flow & 75.99 & 76.36 & 75.98  \\
 BERT-L-CT-SED & 76.92 & 76.96 & 76.77 \\
 BERT-L-CT-SED-Flow & 77.26 & 77.15 & 76.48 \\
 \hline
   \textbf{Using NLI data} &  & & \\
 \hline
 BERT-L-NLI & 73.32 & 74.12 & 73.5 \\
 BERT-L-NLI-Flow & 76.42 & 76.6 & 76.32 \\
 BERT-L-NLI-CT & 77.57 & 77.96 & 77.95 \\
 BERT-L-NLI-CT-Flow & 77.54 & 77.75 & 77.69  \\
 BERT-L-NLI-CT-SED & 77.42 & 77.91 & \B 78.42 \\
 BERT-L-NLI-CT-SED-Flow & \B 77.94 & \B 78.05 & 78.27  \\
  \hline
\end{tabular}
\end{adjustbox}
\label{LayerPoolTable}
\end{table}

\section{Hyperparameters}
\label{Hyperparameters}

\subsection{Siamese Natural Language Inference (NLI)}
All NLI models are trained according to the parameters and settings given in the original paper of \citet{SBert}. Additionally, the code used for all NLI models is from the official S-BERT Github\footnote{\UrlFont{github.com/UKPLab/sentence-transformers}}. The labeled data used is the combined datasets of SNLI \citep{SNLI} and MultiNLI \citep{MultiNLI}, which results in 1 million labeled training examples. We perform one epoch over the data using the Adam Optimizer, a learning rate of $2e^{-5}$ and a linear warm-up schedule over $10\%$ of the data.

\subsection{Contrastive Tension (CT)}
All CT models are trained according to the parameters given in the original paper of \citet{carlsson2021semantic}. Additionally, the code used for all CT models is from the official CT Github: {\scriptsize \texttt{github.com/FreddeFrallan/Contrastive-Tension}}. Using the RMSProp optimizer \citep{RMSProp} with a fixed learning rate schedule that decreases from $1e^{-5}$ to $2e^{-6}$, we perform 50,000 update steps. Sentences randomly sampled from the English Wikipedia\footnote{We use the 20200820 Wikipedia-dump.},
where we have $7$ nonidentical sentence pairs for each positive sentence pair and a batch of 16. Hence, every batch consists of $2$ positive sentence pairs and $14$ negative sentence pairs. 

\subsection{Normalizing Flows (Flow)}
All Flow normalization are trained according to the parameters given in the original paper of \citet{BertFlow}.
Additionally, the code used for all Flow models is from the official BERT-Flow Github\footnote{\UrlFont{github.com/bohanli/BERT-flow}}. The Flow normalization is always done in regards to the combined data from all zero-shot STS tasks, with a learning rate of $1e^{-3}$ and $1$ epoch. The coupling architecture used is described in Table-7 of the original paper of \cite{BertFlow}.

\end{document}